\documentclass[letterpaper, 10 pt, conference]{ieeeconf}  

\IEEEoverridecommandlockouts                              

\usepackage{amsmath} 
\usepackage{amssymb}  
\usepackage{glossaries}
\usepackage{algpseudocode}
\usepackage{graphicx}
\usepackage{algorithm}
\usepackage{algpseudocode}
\usepackage{cite}
\usepackage{booktabs}
\usepackage{multirow}
\usepackage{float}
\usepackage[caption=false]{subfig}

\DeclareMathOperator{\atantwo}{atan2}

\newacronym{eta}{ETA}{Estimated Time of Arrival}
\newacronym{evtol}{eVTOL}{electric vertical takeoff and landing}
\newacronym{fmt}{FMT*}{Fast Marching Tree}
\newacronym{nfz}{NFZ}{no-fly zone}
\newacronym{ov}{OV}{Operational Volume}
\newacronym{prm}{PRM*}{Probabilistic Road Map}
\newacronym{rrt}{RRT*}{Rapidly Exploring Random Trees}
\newacronym{uam}{UAM}{Urban Air Mobility}
\newacronym{uav}{UAV}{Unmanned Aerial Vehicle}
\newacronym{uas}{UAS}{Unmanned Aircraft System}

\title{\LARGE \bf
One-Shot Strategically Deconflicted Route and Operational Volume Generation for Urban Air Mobility Operations
}

\author{Ellis L. Thompson$^{1}$, Yan Xu$^{2}$ and Peng Wei$^{3}$
\thanks{$^{1}$Ellis L. Thompson is with the School of Engineering and Applied Science,
        George Washington University, Washington, DC 20052, USA
        {\tt\small thompson\_e@gwu.edu}}%
\thanks{$^{2}$Yan Xu is with the Centre for Autonomous and Cyberphysical Systems, Cranfield University, Bedford, MK43 0AL, UK
        {\tt\small yanxu@cranfield.ac.uk}}%
\thanks{$^{3}$Peng Wei is with the School of Engineering and Applied Science,
        George Washington University, Washington, DC 20052, USA
        {\tt\small pwei@gwu.edu}}%
}

\begin{document}

\maketitle
\thispagestyle{empty}
\pagestyle{empty}

\begin{abstract}

In the UAM space, strategic deconfliction provides an all-essential layer to airspace automation by providing safe, pre-emptive deconfliction or assignment of airspace resources to airspace users pre-flight. Strategic deconfliction approaches provide an elegant solution to pre-flight deconfliction operations. This overall creates safer and more efficient airspace and reduces the workload on controllers. In this research, we propose a method that constructs routes between start and end nodes in airspace, assigns a contract of operational volumes (OVs) and ensures that these OVs are sufficiently deconflicted against static no-fly zones and OVs of other airspace users. Our approach uses the A* optimal cost path algorithm to generate the shortest routes between the origin and destination. We present a method for generating OVs based on the distribution of aircraft positions from simulated flights; volumes are constructed such that this distribution is conservatively described.

\end{abstract}

\section{Introduction}
\label{sec:introduction}
Safe and efficient flight planning is fundamental in air operations to ensure the safety of operators within the airspace. Additionally, efficient planning effectively uses the available airspace to provide optimal traffic throughput, reducing the burden of work for controllers in an environment with ever-increasing demand. Both the European Aviation Safety Authority and the Federal Aviation Administration predict a surge in \glspl{uas} operations with established numbers of both \glspl{uas} and \gls{evtol} vehicles by 2025 and 2024 respectively\cite{easaUAM,faaforecast}.

To cope with this foreseen demand, the aviation sector is moving toward a more autonomous ecosystem. Where current air traffic control techniques rely on pilot-to-controller communication, the \gls{uam} concepts favour intelligent and autonomous techniques. While tactical deconfliction is an obvious integration point en route, efficient strategic planning can ensure less congested airspace and reduce reliance on tactical methods.

In our previous work\cite{thompson2023}, we proposed a framework for route generation along with the generation of accompanying \glspl{ov}. The results showed a valid candidate path could be generated through an airspace of both static and dynamic airspace constraints. Following the route generation, contracts of \glspl{ov} could then be generated, with the addition of a black box simulator to simulate aircraft movements and leveraging the reachability analysis tool: DryVR \cite{dryvr}. 

In this paper, we build on our previous work by introducing a new \gls{ov} construction technique and adapting an A* based algorithm for path planning. To ensure this new approach improved on the previous system, we generated the following desideratum:

\begin{enumerate}
    \item Contracts of \glspl{ov}, and by extension their accompanying routes, should be deconflicted with both static \glspl{nfz} and \glspl{ov} from proposed contracts belonging to other airspace agents.
    \item There should be an improved synergy between the route generation and \gls{ov} generation, with the former taking precautionary measures to ensure minimal potential overlap between \glspl{ov} and, by extension, reduce route reconstruction frequency.
    \item The route generation algorithm should prioritise optimal route generation, reducing the distance travelled. The existing \gls{rrt} solution should be reconsidered to do this.
    \item The system should provide a candidate route and accompanying contract of \glspl{ov} in a \textit{one-shot} manner. A \textit{one-shot} approach will aid in reducing the ``plan, check, re-plan" reducing the time spent planning a flight pre-departure.
\end{enumerate}

Additionally, we explore alternative approaches to generating the \gls{ov} bounds. We aimed to improve the tightness of the generated \glspl{ov} around the route while omitting any potential adverse side effects as a result of the wrapping effect \cite{Neumaier1993}, the volume grows exponentially in size becoming non-descriptive and overly conservative. We also refer to the candidate route and contract as being generated in a \textit{one-shot} manner. In this project, we describe the system as \textit{one-shot} when: exactly one candidate route and accompanying contract is generated, which is deconflicted with all \glspl{nfz} and other contracts, without the need for several iterations. More precisely, the generated route is optimal and deconflicted with all \glspl{nfz} and other contracts such that the resultant \glspl{ov} are also deconflicted, so no adjustments are needed.

The structure of the remainder of this paper is as follows: Section \ref{sec:related work} provides an overview of related work. Fundamental concepts and formal definitions are presented in Section \ref{sec:concepts}. Section \ref{sec:route generation} covers the approach to route generation and the preliminary route-\gls{ov} deconfliction. In Section \ref{sec:OV generation}, the approach used to generate \glspl{ov} is presented. Experimental results and comparisons to our previous framework\cite{thompson2023} are given in Section \ref{sec: results}. Finally, a discussion and concluding remarks are presented in Section \ref{sec:conclusion}.

\section{Related Work}
\label{sec:related work}

\subsection{Route Generation}
There are several approaches to route generation, for the movement of \glspl{uav}. Our previous work examined techniques using \gls{rrt} as a conceptual baseline\cite{rrt}. Our approach builds on the finite node concept imposed by\cite{rrtfn} to improve computation time and memory storage space. A \textit{rope pull} approach described in\cite{rrtrope} optimises the generated route, removing excess turns or bends through simulating a \textit{rope pull}. In essence, a route generated using \gls{rrt} could be optimised by directly connecting points that have a line of sight to another. Like \gls{rrt} based approaches \gls{prm}\cite{prm}, takes random samples from the configure space. Unlike \gls{rrt}, \gls{prm} does not find a path but generates a practical graph for navigating through a region; a query phase then uses a path-finding algorithm to search for a valid route.

For \gls{rrt} and \gls{prm}, the random nature of the node addition prevents these from having a guaranteed optimal solution. Classical approaches to optimal path planning include: Dijkstra's algorithm\cite{Dijkstra1959}, A*\cite{Hart1968} and the Bellman-Ford algorithm\cite{Ford1956,Bellman1958}, of which all produce minimum-cost paths. For generating optimal paths in a dynamic environment where the obstacles can change, D* and D*-Lite\cite{Stentz1994,dstarlite} provide optimal and efficient solutions for these partially known environments.

\subsection{Reachability Analysis}
Reachability analysis is the computational process of computing reachable states from a given initial state. DryVR\cite{dryvr}, is a data-driven approach to verification for automotive systems. Part of this approach is to perform reachability analysis and generate reachtubes using a black-box simulator. The bounded reach ability algorithm uses learned sensitivities from the trajectory data to overestimate the trajectories provided by the simulator.

Other approaches also focus on Lagrangian methods or an analytical solution\cite{ctrl,Gruenbacher_2020,gotube}, the latter of which can perform slower but is more robust than sensitivity and discrepancy approaches, and the former potentially provides a solution not as susceptible to the wrapping effect\cite{Neumaier1993}. Another common approach is that of bloating techniques \cite{Maidens2015,Fan2017}. The area is artificially increased in size until some parameter is met and exists in many reachability analysis approaches to extend the bounds of the generated volumes.

\subsection{Strategic Planning}
Strategic Deconfliction\cite{Sacharny2022} approaches usually focus on pre-flight stratagem intending to either improve airspace utilisation or improve the likelihood of decreased en route conflict states between aircraft agents. Generating routes that have been strategically deconflicted with other proposed routes, as in\cite{bertram2020scalable, chaimatanan:hal-00868450,Tang2016,Berling2017}, can make efficient use of the available airspace, reducing the number of en route conflict states through adapting routes around areas of high concentrated traffic flow.

\section{Fundamentals}
\label{sec:concepts}

\subsection{Contracts and Operational Volumes}
An \gls{ov} is a finite 4-dimensional region of airspace. The \gls{ov} describes where an aircraft is expected to operate within a limited duration. More formally, we adapt the definition provided in\cite{SkyTrakx}; For the airspace, $\mathcal{X} \subseteq \mathbb{R}^3$, an \gls{ov} can be described as a tuple, seen in Equation \ref{eq: ov tuple}, where $R_k\subseteq\mathcal{X}$ and $T_k$ is monotonically increasing time.
\begin{equation}
    \mathcal{C} = (R_1,T_1),(R_2,T_2),...,(R_n,T_n)
    \label{eq: ov tuple}
\end{equation}

Each \gls{ov} ($\mathcal{C}$) exists for a finite time $\mathcal{C}^{dur} = T_n-T_1$. To obtain the total volume over $t_d$, we can take the union of the \gls{ov} over each interval:
\begin{equation}
    \mathcal{C}^{vol} = \bigcup_{k=1}^{n}R_k
\end{equation}

A contract ($\Xi$), by extension, is the collection of \glspl{ov} over a route. The duration of the contract ($\Xi^{dur}$) is therefore $\sum^N_{i=1}C^{dur}$ for all \glspl{ov} in the contract.

In\cite{thompson2023}, we extended the definition of an \gls{ov} to match that in Equation \ref{eq: extended ov def}, adding an additional element $D$, which describes the distribution of the aircraft within the \gls{ov} at time $t$. The approach to this previously represented the area of the \gls{ov} as a set of grid squares and assigned a value, or probability, of an aircraft present in the square. This approach obfuscated the shape of the recently generated \gls{ov} and was needlessly complex. In Section \ref{sec:building ovs from sim}, we provide a detailed description of how we have redefined $D_n$ for this work.

\begin{equation}
    \mathcal{C} = (R_1,T_1,D_1),(R_2,T_2,D_2),...(R_n,T_n,D_n)
    \label{eq: extended ov def}
\end{equation}

\subsection{A* Search Algorithm}
The A* search algorithm\cite{Hart1968} is a graph traversal and path search algorithm with a worst-case complexity of $O(b^d)$, where $b$ is the branching factor and $d$ is the depth of the desired goal. It is heuristically driven, prioritising exploring nodes that minimise $f(n)=g(n)+h(n)$, where $g(n)$ is the actual distance from the start node ($\Gamma_{start}$) to the current node ($\Gamma_n$) and $h(n)$ is the heuristic value, commonly the direct distance from $\Gamma_n$ to the goal node $\Gamma_{goal}$.

\section{Route Generation}
\label{sec:route generation}
For route generation, we employed the A* algorithm. Common in route planning tasks for its guaranteed optimality, we also have the added benefit of predicting where the aircraft could be when given a departure time and cruise speed. This benefit, especially over our previous approach using \gls{rrt}, allows us to perform preliminary route-\gls{ov} deconfliction based on this predicted position. By performing this preliminary route-\gls{ov} deconfliction, we can significantly increase the system's overall performance as we can expect fewer inter-\gls{ov} conflicts when generating them. 

\subsection{Point Expansion}
The environment in which we apply our A* algorithm is not constrained to a grid. We sacrifice the improved efficiency that comes with a predetermined grid of nodes in favour of an open environment, reducing the overall memory taxation of the search.

\begin{figure}[ht]
    \centering
    \includegraphics[width=0.7\linewidth]{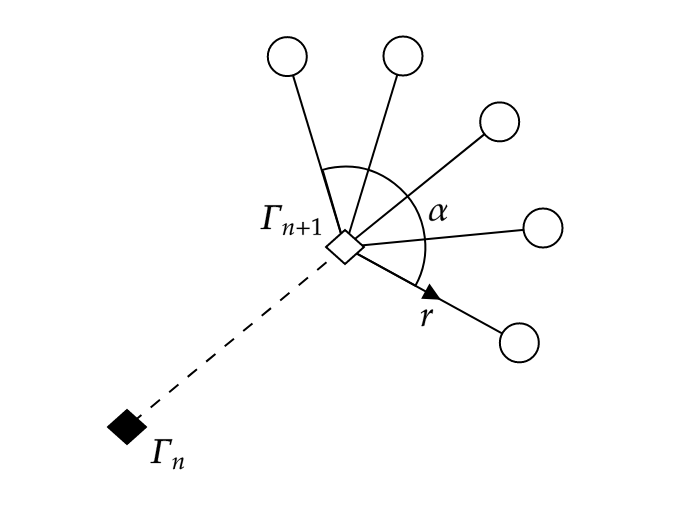}
    \caption{New nodes are generated at regular intervals along the arc of angle $\alpha$ where the centre of the arc extends the path $\Gamma_n \rightarrow \Gamma_{n+1}$.}
    \label{fig:node extension}
\end{figure}

We build a graph during the node expansion phase, providing the nodes introduced follow some rules. Firstly, to reduce the number of sharp angled turns, we limit the nodes to only be generated in a finite arc of radius $r$ and angle $\alpha$ such that the centre of the arc extends the line $\Gamma_n \rightarrow \Gamma_{n+1}$ as shown in Figure \ref{fig:node extension}. To reduce the search space, inserted nodes within some bounded distance of nodes already explored in the environment are counted as the same node. If this is the case, then the $g(n)$ values are compared, and if the inserted node has a lower $g(n)$ value, then the explored value is updated and re-expanded to ensure optimality in the route.

For a given previously explored point, the heuristic also needs to be calculated once. As outlined in Equation \ref{equ: heuristic}, the heuristic function does not include any information about previously visited nodes. With the heuristic function being this approach's most considerable computational burden, the added benefit of only calculating a heuristic once significantly improves computation speed.

During the expansion phase, nodes are categorised as either \textit{valid} or \textit{invalid}. A \textit{valid} node is one which exists within the constraints of the airspace and a valid connection to a parent node exists. A valid connection is similarly defined as a connection that does not violate the constraints of the airspace or intersect with any \gls{nfz}. A node with an invalid connection is, therefore, a node with a path to its parent node that violates the constraints of either the airspace or a \gls{nfz}. To expand further, an \textit{invalid node} is defined as any node within or too close to the region of a \gls{nfz} or outside the operational area bounds. Nodes with invalid connections are handled differently; rather than being excluded from the expansion, they are immediately placed into the \textit{closed} list. This allows us to pre-calculate and store the heuristic, reducing the number of calculations, if the node is ever reached again via a safe route.

\subsection{Heuristic Function}
The heuristic function used in the A* algorithm adapts the approach proposed in\cite{Razzaq2018}. This method first calculates the angle from the node ($\Gamma_n$) to all vertices on the polygon $\mathcal{Z}_j$ as illustrated in Equation \ref{equ: angles on a polygon}. The \gls{nfz} polygons observed are primarily concave, providing a challenge for the route-finding algorithm.
\begin{equation}
    \Theta(\Gamma_n,\mathcal{Z}_j) = \{\theta(\Gamma_n, \mathcal{Z}^i_j) - \theta(\Gamma_n,\Gamma_{goal}), \forall i \in \mathcal{Z}_j\}
    \label{equ: angles on a polygon}
\end{equation}

\begin{figure}[ht]
    \centering
    \includegraphics[width=0.7\linewidth]{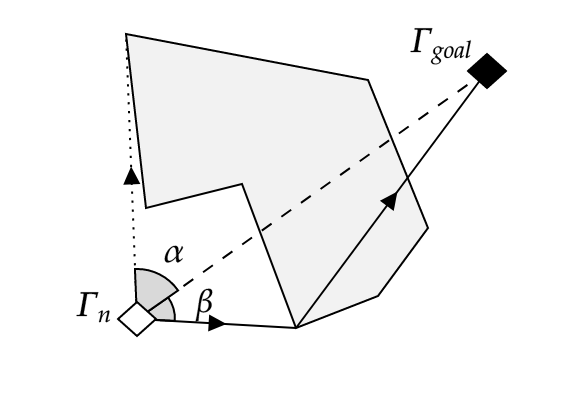}
    \caption{The heuristic for a point whose line of sight is obscured by a \gls{nfz} is the distance to the extreme point on that polygon with the smallest angle and then the distance to the goal. This is represented by the path $\beta$, and the line $\alpha$ is not used as the distance $\theta_\alpha>\theta_\beta$.}
    \label{fig:heuristic dem}
\end{figure}

This will yield a positive value if the point lies to the right of the direct line $\Gamma_n \rightarrow \Gamma_{goal}$ and a negative value otherwise. The extremes are then found from this set, that is, the point furthest left and right of the line $\Gamma_n \rightarrow \Gamma_{goal}$, this can be obtained by finding the minimum and maximum values in $\Theta(\Gamma_n,\mathcal{Z}_j)$. The heuristic, presented in Equation \ref{equ: heuristic} and demonstrated in Figure \ref{fig:heuristic dem}, is then the distance between $\Gamma_n$ and the extreme vertex $\mathcal{Z}_j^i$, whose absolute angle is the smallest from the line $\Gamma_n \rightarrow \Gamma_{goal}$ and the distance from $\mathcal{Z}_j^i$ to $\Gamma_{goal}$. We additionally multiply this by $\omega$ where $1.1 \leq \omega \leq 1.5$; which improves the performance of the A* algorithm in favouring exploring points closer to the goal.

\begin{equation}
    h() = \omega(d(\Gamma_n, \mathcal{Z}^i_j)+ d(\mathcal{Z}^i_j, \Gamma_{goal}))
\label{equ: heuristic}
\end{equation}

\subsection{Route-\gls{ov} Deconfliction}
We can predict the time an aircraft reaches a node ($\Gamma_n$) through the distance travelled from $\Gamma_{start}$ and providing the cruise speed $V_{cruise}$, that is, the expected speed of the aircraft. In this approach, we assume that the route is only generated for the cruise phase of the flight, so any departure or arrival altitude and speed profiles are not considered. To further perform route-\gls{ov} deconfliction, the system must store a representation of the airspace at discreet time intervals.

The airspace is divided into time bins of duration $t_d$ for a given operational area. \glspl{ov} are then placed into these brackets based on the start time and perceived end time of the \gls{ov}. This means that an \gls{ov} can exist in multiple time bins if either the $t_d$ of the bin is shorter than that of the \gls{ov} or the \gls{ov} lies near the boundary of a time bin. This airspace binning reduces the search scope to only \glspl{ov} that lie near the node temporally.

Suppose a node lies within the bounds, both spatially and temporally, as an \gls{ov} or the path between nodes intersects an \gls{ov}, again both spatially and temporally, the node is treated as having an invalid connection. The deconfliction then occurs by observing a pair of two points, $(\Gamma_n, \Gamma_{n-1})$ that is the node to add and its parent, and the connecting line between them. The respective \glspl{eta} of both nodes are then compared with the time bins in the airspace, and \glspl{nfz} in bins that overlap the \glspl{eta} are obtained. The process remains the same as deconflicting with \glspl{nfz}. As is recalled from handling nodes with invalid connections for \glspl{nfz}, the node is still processed but added directly to the \textit{closed} list of the A* algorithm.

This approach prevents nodes, or routes between nodes, from directly conflicting with \glspl{ov} already proposed in the airspace. We can then further decrease the possibility of an \gls{ov} generated along the route conflicting with the existing \glspl{ov} by imposing a distance constraint on the path between nodes $(\Gamma_n, \Gamma_{n-1})$ and any temporally aligned \glspl{ov}.

We can further add to the deconfliction approach by considering the distribution of aircraft with an \gls{ov}. As seen in the next section, we can represent the aircraft's position within an \gls{ov} as a density map at any given time. We expect, within an \gls{ov}, that aircraft will travel closer to the centre with fewer outliers towards the edges of the \gls{ov}. Considering this distribution, the probability of observing an aircraft along a proposed connection can be calculated when an overlap is observed. This can be achieved by evaluating the probability density function for the points along the connection $(\Gamma_{n-1}, \Gamma_n)$ of the multivariate normal distribution that can be obtained from the \gls{ov}. We can then add this to the heuristic function in Equation \ref{equ: new heuristic} where $\beta \geq 0$ is some constant and $\delta(\Gamma_{n-1}, \Gamma_n)$ is the maximum probability of conflict expanded in Equation \ref{equ:delta}. 

\begin{equation}
    h() = \omega(d(\Gamma_n,\mathcal{Z}_j^i)+d(\mathcal{Z}_j^i, \Gamma_{goal})) + \beta \delta(\Gamma_{n-1}, \Gamma_n)
    \label{equ: new heuristic}
\end{equation}
\begin{equation}
    \delta(\Gamma_{n-1}, \Gamma_n) = \max_{\xi\in\Xi}\max_{\mathcal{C}\in\xi}P((\Gamma_{n-1}, \Gamma_n),\mathcal{C})
    \label{equ:delta}
\end{equation}

\section{Generating Operational Volumes}
\label{sec:OV generation}

In our previous approach, \gls{ov} generation was performed using a combination of a simulator, namely BlueSky\cite{joost2016}, and the reachability analysis tool DryVR\cite{dryvr}. The DryVR framework performs reachability analysis by learning a global discrepancy function based on the data provided by the simulation. The resultant reachtubes are 3-dimensional bounds between regular, discreet time intervals which we adopted to build the \glspl{ov} influenced by the approach outlined in\cite{SkyTrakx}. However, this approach is sensitive to the data distribution and suffers from the \textit{wrapping effect}\cite{Neumaier1993}. We propose a new approach, describing the data distribution as ellipsoids for discrete time intervals. The ellipsoids are based on the distribution of the recorded aircraft.

\subsection{Utilising the Simulation}
\label{sec: using the sim}
As with the approach referenced in\cite{dryvr,SkyTrakx}, our approach uses simulation data to build the bounding spheroids. To generate this data BlueSky\cite{joost2016} is used again for its variety of aircraft models along with its Python implementation, making fine control of the simulation and data extraction trivial and well-integrated with our architecture. 

Once a route has been generated using the method from Section \ref{sec:route generation}, the set of waypoints ($\mathcal{W}$) are generated for BlueSky. Some initial uncertainty is also injected into the system, and $N$ aircraft are initialised in a radius around the initial waypoint $w_0$. The simulation then proceeds for the duration $t_d$, recording the state of the aircraft at regular intervals. The state of the airspace at any given time is denoted as $S_t^N$; therefore, the state of an individual aircraft at $t$ is defined as $S_t^n, n\in N$. The state information stored, presented in Equation \ref{equ: state}, includes the positional data, altitude, active waypoint ($w\in \mathcal{W}$) and the airspeed of the vehicle ($V$).
\begin{equation}
    S_t^N=(lat^N_t,lon^N_t, alt^N_t, w_t^N, V_t^N)
\label{equ: state}
\end{equation}

Once the simulation has run for the duration $t_d$, new uncertainty is injected into the system. This is achieved by generating new aircraft proportionally relative to active waypoints and the number of aircraft approaching them. This approach naturally accounts for the spread of the aircraft in the simulation occurring from uncertainty in the initial position and external effects such as wind. To demonstrate this approach, if there are recorded three active waypoints out of the $N=100$ aircraft in the environment $(w_\alpha, w_\beta, w_\gamma)\in \mathcal{W}$, with the distributions as $(25,71,4)$, then simulation run to $t_d$ will start with new aircraft approaching the aforementioned waypoints following the distribution $(w_\alpha=25, w_\beta=71,w_\gamma=4)$ respectively.

\subsection{Building \glspl{ov} from Simulation Data}
\label{sec:building ovs from sim}
After each simulation has been completed to $t_d$, in parallel to the new aircraft generation, the \gls{ov} for the previously completed run is generated. The data passed to the \gls{ov} generator is in the form of $S_t^N$, firstly a sample of this data is taken $S_t^M \subset S_t^N$ such that $|S_t^N-S_t^M| \leq |S_t^M|$ and $|S_t^M| \geq15$. This separates the state space into two subsets $S_t^M$ is used to build the \gls{ov}, and the remainder is later used to validate the \gls{ov} includes sufficient points. The samples are drawn at random.

In this approach, we want to introduce a method that can both describe the area that the \gls{ov} will occupy while simultaneously describing the distribution of these aircraft. We construct a covariance matrix ($\Sigma_t$) from the state $S_t^M$, specifically over the latitude, longitude and altitude components and obtain the mean position $\mu_t$. Next, the eigenvectors $v_i$ and eigenvalues $\lambda_i$ are obtained.

The eigenvalues ($\lambda_i$) are used to derive the width and height of the resulting ellipse using Equation \ref{eq: ellipse size} where $l_w$ and $l_h$ are the lengths of the major and semi-major axes, respectively. The value $z$ is the confidence interval for a normal distribution, described in Equation \ref{eq: ellipse inclusion} where $x$ is the desired minimum inclusion, $PPF$ is the per cent point function for the normal distribution and $\alpha$ is a constant to bloat the ellipse artificially.

\begin{equation}
    l_{w,h} = 2z\sqrt{\lambda_{w,h}}
    \label{eq: ellipse size}
\end{equation}
\begin{equation}
   z = \left| PPF\left(\frac{1-x}{2}\right)\right|+\alpha
   \label{eq: ellipse inclusion}
\end{equation}

Finally, the ellipse's rotation angle is calculated from the eigenvectors $v_i$. It should be noted that for a $2\times 2$ covariance matrix, two eigenvectors are obtained, each containing two values where $v_1$ and $v_2$ match with the eigenvalues $\lambda_1$ and $\lambda_2$ respectively. To obtain the rotation, we use the 2-argument arctangent: $\theta = \atantwo(v_{1,0}, v_{0,0})$.

This generates a single ellipse $R_i$ for $T_i$ as described in our \gls{ov} definition in Section \ref{sec:concepts}. We use the resultant set $S = S_t^N-S_t^M$ and our desired minimum inclusion variable $x$ to verify that the ellipse is sufficiently encompassing. To verify, we calculate the Mahalanobis Distance\cite{mahalanobis1936generalized}, Equation \ref{eq:mDist}, first across $S_t^M$ and then set the $x^{th}$ percentile distance as the threshold distance.

\begin{equation}
    M_{Dist}(y) = \sqrt{(y-\mu_t)\Sigma_t^{-1}(y-\mu_t)'}
    \label{eq:mDist}
\end{equation}

Then the Mahalanobis Distance is calculated for all values in $S$ and it is asserted that:
\begin{equation*}
    \left(\frac{1}{|S|}\sum^{|S|}_{i=1}\left[M_{Dist}(S_i) <= Threshold\right] \right)>= x
\end{equation*}

If it is found that the number of points included falls below the threshold $x$ the bloating factor $\alpha$ can be increased until the threshold is met. This is done in line with the generation of the ellipse.

As these ellipses are generated, they can be added to the \gls{ov}, such that $D_n$ consists of the covariance matrix $\Sigma$ and the value $z$, $T_n$ is set as the start time of the interval, and $R_n$ while not stored can be derived as shown in this section. Repeating this process over each time interval forms a single \gls{ov} of length $\mathcal{C}^{dur}$.

\section{Experimental Results}
\label{sec: results}

\begin{figure}[ht]
    \centering
    \includegraphics[width=0.8\linewidth]{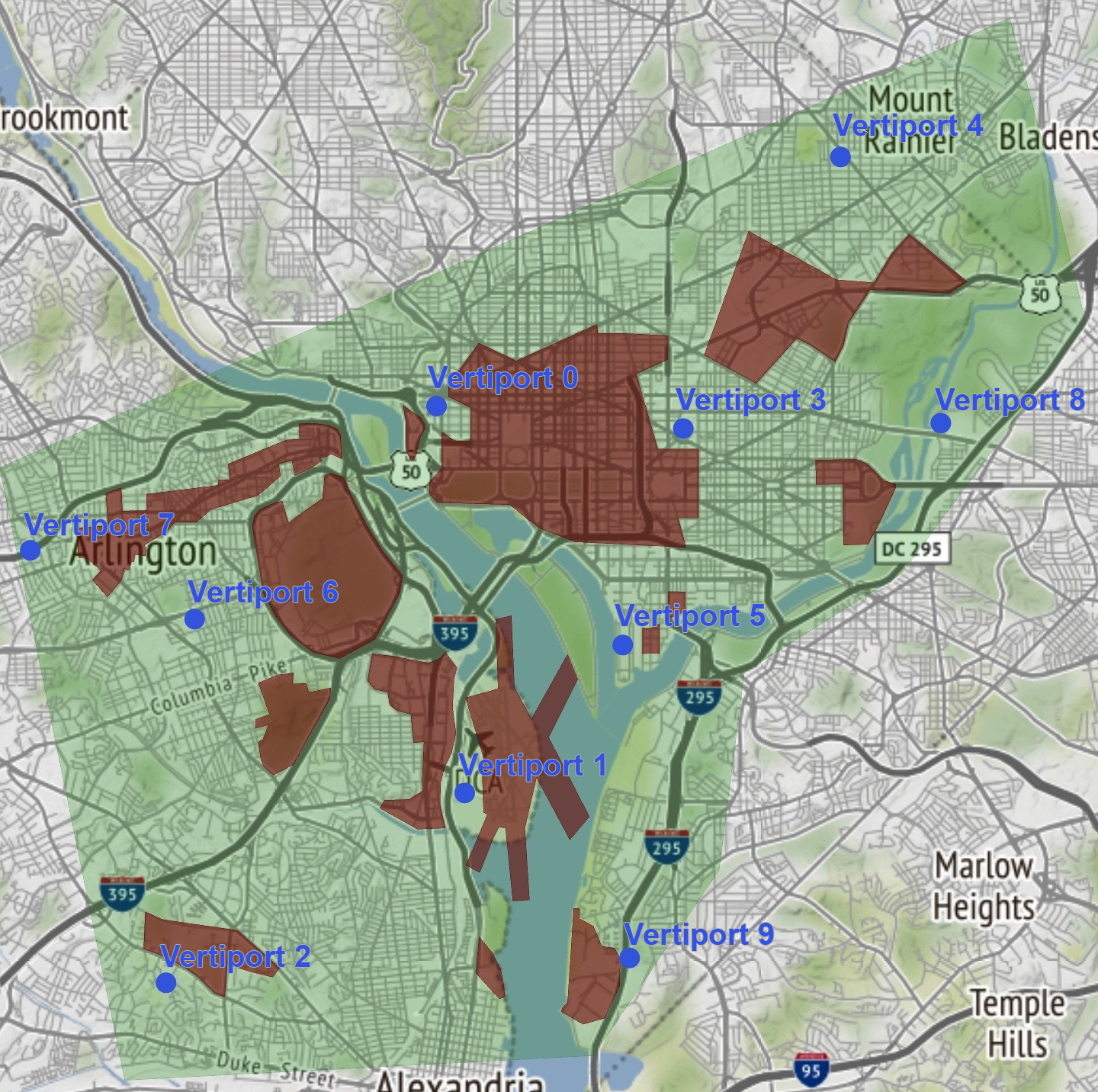}
    \caption{The mock airspace used for testing the systems. It consists of 10 arrival/departure nodes (yellow pins) and concave \glspl{nfz} (red polygons), designed to challenge the route-finding algorithm. The green lines denote the operational bounds of the airspace.}
    \label{fig: airspace}
\end{figure}

A mock airspace environment was created to test the system, as shown in Figure \ref{fig: airspace}. The airspace consists of ten arrival and departure nodes resulting in 45 unique routes. Additionally, to stress the route-finding algorithm and heuristic, the \glspl{nfz} presented are concave, proposing a challenging environment for path finding algorithms.

The experiments are compared against our previous approach described in \cite{thompson2023}. We first explore the route generation capabilities, examining route construction time and the resulting lengths of the routes generated. The \gls{ov} generation is then examined. We compare the \gls{ov} generation times collectively of both approaches as well as the overall volume of the generated \glspl{ov} and their relative accuracy. These experiments are done in an environment with no other traffic, so deconfliction is not evaluated in these initial experiments. The final experiment stresses the proposed approach in an environment with other existing contracts, evaluating the ability to generate and deconflict routes and the accompanying contract of \glspl{ov}. All experiments are run using the provided Amazon drone in the BlueSky environment as part of the OpenAP \cite{sun2020openap} dataset. All experiments were run on a 2020 MacBook pro using an Intel 2GHz i5 processor and 32GB of memory.


\subsection{Route Generation}
For this experiment, both the \gls{rrt}-based route generation method from \cite{thompson2023} and the A* method proposed in this paper are tasked with finding routes within the environment shown in Figure \ref{fig: airspace}. We initially hypothesise that the A* method, being a cost-optimal algorithm, would be able to find shorter paths than the \gls{rrt}-based solution. 

\begin{figure}[ht]
\subfloat[The runtimes of the \gls{rrt}-based method for the five test routes.\label{fig:rrtrime}]{\includegraphics[width=\linewidth]{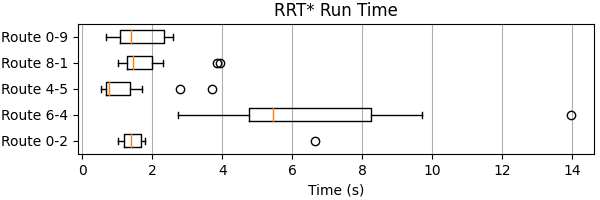}}\qquad
\subfloat[The distances of the\gls{rrt}-based method for the five test routes.\label{fig:rrtdistances}]{\includegraphics[width=\linewidth]{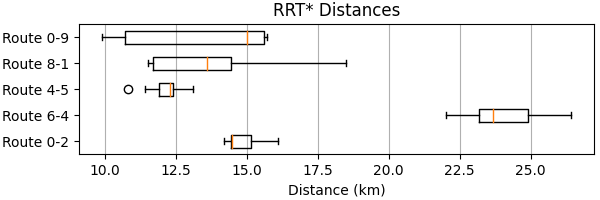}}
\caption{}
\end{figure}

The results in Table \ref{tab:Route distances} demonstrate that, although slower, A* produces significantly shorter routes. Due to the stochastic nature of the \gls{rrt}-based method ten experiments were run for each route and the averages used for comparison. Figures \ref{fig:rrtrime} and \ref{fig:rrtdistances}, show the distribution of both the run times and distances of the \gls{rrt} route planning approach. Not that the run times remain relatively consistent likewise with the route distances over shorter routes. Route 6-4 has the largest variance in runtime. This is likely due to the origin vertiport being between two concave \glspl{nfz}. The stochasticity of the \glspl{rrt} method can be a limiting factor for that method of route generation as the approach adopted returns the first found route without further exploration.

It is worth nothing that, as can be observed in Figure \ref{fig:my_label}, on occasion routes may not always be optimal. Due to the nature of the A* algorithm relying on predetermined nodes, if the initial route is blocked, the resulting route generated may significantly divert from this optimal path. This performance can be improved by decreasing the distance between nodes, as well as the number of angles as visualised in Figure \ref{fig:node extension}. This comes at the impedance of computation time with the resulting routes taking longer to generate.

\begin{table}[ht]
\centering
\caption{Comparison of A* and RRT route generation methods}
\label{tab:Route distances}
\begin{tabular}{@{}cccccc@{}}
\toprule
\multirow{2}{*}{Route} & \multirow{2}{*}{\begin{tabular}[c]{@{}c@{}}Direct \\ Distance (km)\end{tabular}} & \multicolumn{2}{c}{\begin{tabular}[c]{@{}c@{}}Route Dist\\ (km)\end{tabular}} & \multicolumn{2}{c}{\begin{tabular}[c]{@{}c@{}}Generation\\ Time (s)\end{tabular}} \\ \cmidrule(l){3-6} 
                          &                                                                                  & A*                                         & RRT                              & A*                                 & RRT                                          \\ \midrule
0-2                       & 10.01                                                                            & \textbf{12.39}                             & 14.83                            & 10.04                               & \textbf{1.86}                               \\
6-4                       & 12.48                                                                            & \textbf{13.04}                             & 23.98                            & 7.08                                & \textbf{6.74}                               \\
4-5                       & 8.39                                                                             & \textbf{9.06}                              & 12.11                            & 3.89                                & \textbf{1.28}                               \\
8-1                       & 9.48                                                                             & \textbf{10.79}                             & 13.73                            & 13.08                               & \textbf{1.90}                               \\
0-9                       & 9.19                                                                             & \textbf{9.88}                              & 13.43                            & 3.07                                & \textbf{1.61}                               \\ \bottomrule
\end{tabular}
\end{table}

\subsection{\gls{ov} Generation}
In this experiment, we compare the ellipse based \gls{ov} generation presented in this paper against the reachability analysis method DryVR \cite{dryvr}, used in our previous work \cite{thompson2023}. The routes used were pregenerated using the A* route generation method. These were then supplied to both \gls{ov} generation methods.

Both approaches utilise the BlueSky \cite{joost2016} air traffic simulator to generate the data used in \gls{ov} generation. To ensure consistency, the initial starting points of each scenario were identical for each pair of experiments. Additionally, the speeds of the aircraft were uniformly distributed between $13m/s$ and $17m/s$. For the DryVR method, the overlap offset was set to 20 seconds. The results are presented in Table \ref{tab:ov vs dryvr}.

\begin{table}[ht]
\caption{Comparison of our \gls{ov} generation against DryVR}
\label{tab:ov vs dryvr}
\begin{tabular}{ccccccc}
\toprule
\multirow{2}{*}{Route} & \multicolumn{2}{c}{Time (s)} & \multicolumn{2}{c}{\begin{tabular}[c]{@{}c@{}}No.\\ OVs\end{tabular}} & \multicolumn{2}{c}{\begin{tabular}[c]{@{}c@{}}Avg Area\\ (km$^2$)\end{tabular}} \\ \cline{2-7} 
                       & Ours              & DryVR    & Ours                                 & DryVR                          & Ours                                   & DryVR                                  \\ \hline
0-2                    & \textbf{88.01}    & 424.51   & \textbf{15}                          & 27                             & 0.045                                  & \textbf{0.032}                         \\
6-4                    & \textbf{86.04}    & 498.43   & \textbf{16}                          & 28                             & 0.045                                  & \textbf{0.028}                         \\
4-5                    & \textbf{49.19}    & 286.61   & \textbf{11}                          & 20                             & \textbf{0.043}                         & 0.055                                  \\
8-1                    & \textbf{67.53}    & 320.56   & \textbf{13}                          & 23                             & \textbf{0.044}                         & 0.057                                  \\
0-9                    & \textbf{50.62}    & 302.76   & \textbf{12}                          & 21                             & 0.044                                  & \textbf{0.037}                         \\ \bottomrule
\end{tabular}
\end{table}

From Table \ref{tab:ov vs dryvr}, we see that the approach presented in this paper is significantly quicker at generating \glspl{ov} than the previously adopted method. This is due to the relatively straightforward and more optimised approach adopted in this paper over that from DryVR. In this new approach, the calculations can be highly parallelised as the result of one \gls{ov} does not rely on the parameters of the preceding \gls{ov}, only aircraft positional data.

The method for generating reachtubes in DryVR does not account for uncertainty in the initial position. Furthermore, our approach produces significantly fewer \glspl{ov}. This is caused by no offset variable being introduced in this approach. This, in turn, yields strict boundaries between \glspl{ov}, as seen in \cite{SkyTrakx}, which the offset value introduced in \cite{thompson2023}, attempted to address. In this new approach, however, uncertainty in the initial position is included through the method outlined in Section \ref{sec: using the sim}. This naturally generates overlapping \glspl{ov}.

\begin{figure}[ht]
    \centering
    \includegraphics[width=0.65\linewidth]{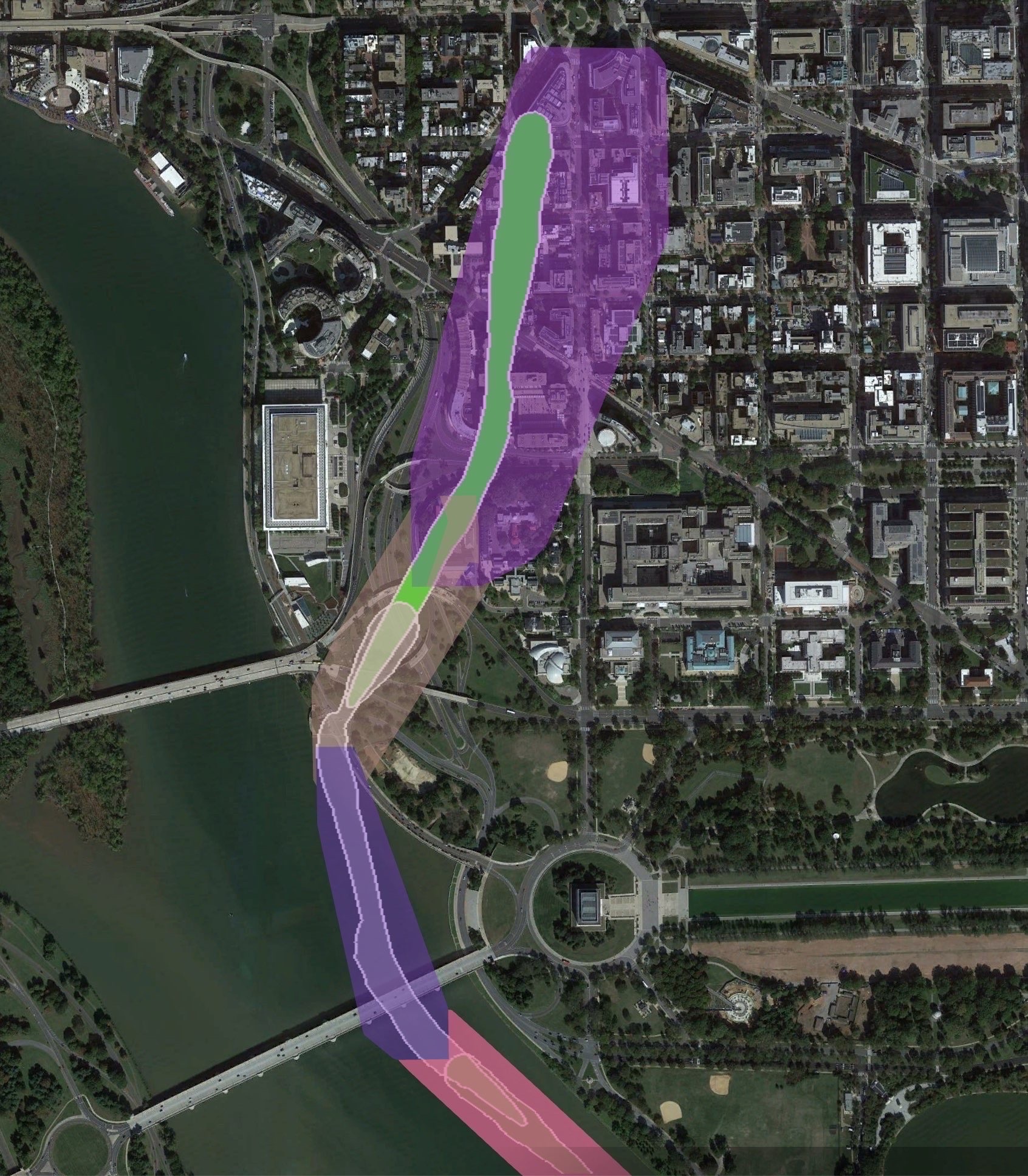}
    \caption{The difference between the \glspl{ov} generated using DryVR \cite{dryvr} and our method can be seen by comparing the shaded regions. The outlined regions are generated using our ellipsoid method and are demonstrated to be tighter laterally than the DryVR approach. The colours are only to aid visibility.}
    \label{fig:ov vs dryvr}
\end{figure}

The final observation is that the average volume of the \glspl{ov} generated using the DryVR approach appears to be smaller. What is not conveyed, however, is how this approach handled initial uncertainty or turns. Figure \ref{fig:ov vs dryvr}, demonstrates the difference in \gls{ov} generation methods, where the ellipsoid \glspl{ov} are outlined in white. The green and purple \glspl{ov} towards the top of the figure is over the initial start point of the route. It can be seen from these two \glspl{ov} that the DryVR method struggles to handle this initial uncertainty, resulting in an \gls{ov} with more conservative bounds when compared to the ellipse-based approach. 

We tested the accuracy by simulating 100 aircraft with added uncertainties to the latitude, longitude and speeds. The aircraft's position was recorded at 1-second intervals and then compared against the generated \glspl{ov}, counting the total points within a valid \gls{ov} for the given time. Both approaches generated contracts with 100\% accuracy across all five routes.

\subsection{Congested Environment}
The system was tested in a high-traffic density environment for the final set of experiments. Flights were progressively created at random over a 5-minute interval. The resulting contracts and departure times were stored and used in the route generation and deconfliction of subsequent flights. This resulted in the environment progressively increasing in density. Additionally, the cruise speeds of the contracts were assigned at random between $13m/s$ and $20m/s$, and to ensure the highest level of complexity, no conflicts between \glspl{ov} of other contracts were allowed. Finally, this experiment was only run with the method proposed in this paper.

This experiment aimed to reach at least 30 contracts in the environment. The system was allowed to generate random start and end node pairs with randomly assigned departure times within the interval. If no route was found, the system would look for the closest departure time, in 30-second intervals, within the 5-minute departure slot. If no route was found, the system would look at a new pair and departure time. This correlates to one departure every 10 seconds.


The departure times were chosen randomly from a Poisson distribution, with the departure and arrival vertiports chosen from a uniform distribution. A second test was run in the BlueSky simulator, which then evaluated conflicts between aircraft following the assigned routes at the previously mentioned departure intervals.

\begin{figure}[ht]
    \centering
    \includegraphics[width=\linewidth]{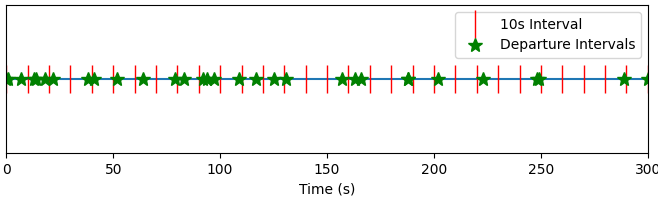}
    \caption{The departure intervals of routes of the 31 contracts. While distributed along the 5-minute window, we observe more departures in the first 2 minutes.}
    \label{fig:deptimes}
\end{figure}

The experiment was run until 31 contracts existed in the environment. Figure \ref{fig:deptimes} shows the departure times over the 5-minute departure window. We observed that more departures were present during the first 2 minutes. This created a denser airspace in those first 2 minutes, resulting in higher computational times, with the average run time being around 3 minutes for both route and \gls{ov} generation.

\begin{figure}[ht]
    \centering
    \includegraphics[width=0.9\linewidth]{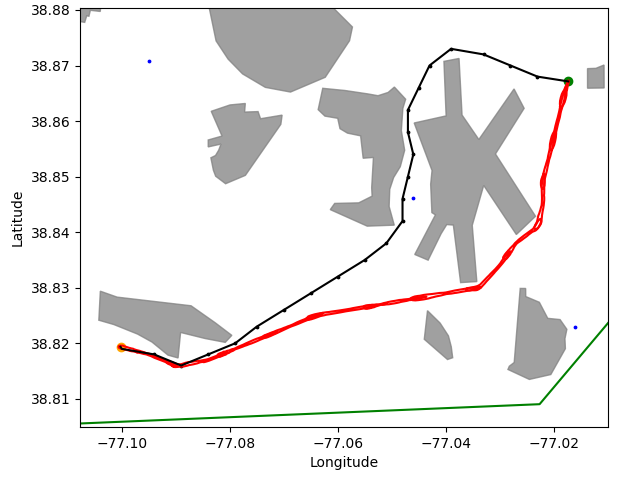}
    \caption{The \glspl{ov} (red) are the first to be generated. This takes the shortest route between the departure and arrival points. The route in black is a deconflicted second route with the opposite departure and arrival points but the same departure time. It has to take an alternative route to remain deconflicted with this existing route and others in the environment (which are not drawn here).}
    \label{fig:my_label}
\end{figure}

When running simulations in BlueSky to verify the number of aircraft existing in confliction contracts \glspl{ov}, we found that all aircraft remained deconflicted with other \glspl{ov}. The further through the experiment, the more challenging the environment became for the system. However, the presented environment provides large open areas with ample space for routing. Overall, the system successfully routed and deconflicted 31 routes within a finite 5-minute departure window. Figure \ref{fig:my_label} shows opposing routes with identical departure times. As the black route was generated later, it takes an alternative shortest route to avoid conflicts with the red \glspl{ov}.

\section{Conclusion}
\label{sec:conclusion}

The research presented in this paper builds on our previous work in \cite{thompson2023}. To build on the previous research, we identified four critical areas of improvement we needed to address to ensure a robust system. 

The first point addressed the need to build contracts where the \glspl{ov} were deconflicted against static \glspl{nfz} along with \glspl{ov} from other contracts. This was addressed primarily in the route generation stage, where connections are deconflicted from the existing airspace obstacles when connecting nodes in the graph. We found the system could generate routes and fully deconflicting contracts of \glspl{ov} in finite airspace. While the system slowed down when more contracts were included, the average full computation time for each route and contract was around 3 minutes. An improvement could arise from a better definition of an airspace manager to dynamically return time-aligned \glspl{ov} to the route builder.

The second point addressed the need for improved synergy between \gls{ov} and route generation. As already identified, the primary deconfliction effort occurs at the route generation stage. From Equation \ref{equ: new heuristic}, new connections are weighted based on the maximum probability of conflict. This is obtained for any given \gls{ov} from the covariance matrix and the centre point. This allows fine control over how much to consider overlaps, potentially expanding $\delta$ further to implement thresholds or minimum separations.

Our previous implementation switched the route generation method to an A* based approach. This satisfies the third requirement and is reinforced by the results in Table \ref{tab:Route distances}, where our A* approach generates shorter routes than the previous \gls{rrt}-based approach. We have also created a heuristic function that efficiently navigates the concave obstacles in the airspace. However, this approach is slower than the \gls{rrt}-based methods and careful parameter optimisation and adjusting the heuristic function could result in faster computational times.

The final point addressed the possibility of designing a one-shot system. We proposed that this approach should only require one iteration for route generation. Without including a re-scheduler, a system that finds a starting time for the flight, the system can generate a route and \glspl{ov} that are deconflicted in a one-shot manner. With a rudimentary scheduler, the system can still run self-contained and successfully generate a route and deconflicted \glspl{ov}, providing one exists. However, if a route is not found initially, then a route will have to be regenerated with a new provided departure time. 

An ideal system would fall between the abilities of A* to produce cost-optimal routes along with runtime efficiency of \gls{rrt}. One middle ground could be derived from an \gls{fmt} based approach \cite{janson2015fast}. Producing probabilistically near-optimal solutions it has been shown to outperform its contemporaries; \gls{rrt} and \gls{prm}. Furthermore, it has been shown a viable option for emergency route planning of commercial aircraft in \cite{aerospace9040180}, lending promise to the viability of a \gls{fmt}-based approach being useful for the planning of \gls{uas} operations. Other classic approaches, such as storing the best common routes, could also significantly improve the computation time of the system at the cost of increased memory usage.
 
In conclusion, we have proposed a one-shot system for generating routes and deconflicted \glspl{ov} for small \gls{uas} vehicles to operate in. We demonstrate that this system, in a random and congested airspace, successfully generates and deconflicts multiple contracts and routes with departure times within a 5-minute interval.

\bibliographystyle{IEEEtran}
\bibliography{IEEEabrv,bibliography}

\end{document}